\pdfoutput=1

\documentclass[11pt]{article}

\usepackage[final]{acl}

\usepackage{times}
\usepackage{latexsym}

\usepackage[T1]{fontenc}

\usepackage[utf8]{inputenc}

\usepackage{microtype}

\usepackage{inconsolata}

\usepackage{graphicx}
\usepackage{caption}
\usepackage{subcaption}
\usepackage{float}
\usepackage{svg}
\usepackage{amsmath} %
\usepackage{lipsum}
\usepackage{colortbl}
\usepackage{booktabs}
\usepackage{nicematrix}
\usepackage[listings]{tcolorbox}
\usepackage[a4paper]{geometry}
\usepackage{changepage}
\usepackage{avm}
\usepackage[utf8]{inputenc}
\usepackage{amsmath}
\usepackage{yhmath}
\usepackage{footmisc}
\newcolumntype{P}[1]{>{\centering\arraybackslash}m{#1}}

\definecolor{mycustomcolor}{HTML}{1F77B4}
\definecolor{nicegreen}{HTML}{3CB043}
\newcommand{\green}[1]{\textcolor{nicegreen}{#1}}
\newcommand{\red}[1]{\textcolor{red}{#1}}
\newcommand{\orange}[1]{\textcolor{orange}{#1}}

\title{Benchmarking Critical Questions Generation: A Challenging Reasoning Task for Large Language Models}

\author{Blanca Calvo Figueras \\
  HiTZ Center - Ixa \\ University of the Basque \\ Country UPV/EHU \\
  \texttt{blanca.calvo@ehu.eus} \\\And
  Rodrigo Agerri \\
  HiTZ Center - Ixa \\ University of the Basque \\ Country UPV/EHU \\
  \texttt{rodrigo.agerri@ehu.eus} \\}

\begin{document}
\maketitle
\begin{abstract}

The task of Critical Questions Generation (CQs-Gen) aims to foster critical thinking by enabling systems to generate questions that expose underlying assumptions and challenge the validity of argumentative reasoning structures. Despite growing interest in this area, progress has been hindered by the lack of suitable datasets and automatic evaluation standards. This paper presents a comprehensive approach to support the development and benchmarking of systems for this task. We construct the first large-scale dataset including $~$5K manually annotated questions. We also investigate automatic evaluation methods and propose reference-based techniques as the strategy that best correlates with human judgments. Our zero-shot evaluation of 11 LLMs establishes a strong baseline while showcasing the difficulty of the task. Data and code\footnote{\url{https://github.com/hitz-zentroa/Benchmarking_CQs-Gen}} plus a public leaderboard are provided to encourage further research, not only in terms of model performance, but also to explore the practical benefits of CQs-Gen for both automated reasoning and human critical thinking.
\end{abstract}

\section{Introduction}

Recent research has documented significant pedagogical concerns regarding the proliferation of Artificial Intelligence conversational interfaces. Specifically, it suggests that these systems might be reinforcing superficial learning processes while simultaneously diminishing users' capacities for critical thinking \cite{hadi_mogavi_chatgpt_2024}.

\begin{figure}[h!]
\footnotesize  
\noindent\rule[0.5ex]{\linewidth}{1pt}

\vspace{5pt}
\begin{subfigure}{1\linewidth}

\raggedright \textbf{Argument}:

\vspace{5pt}

\centering \textit{"I did not use sunscreen yesterday and I was fine, so I don't think you need it today."}  %

\vspace{5pt}

\raggedright \textbf{Critical Questions}:

\begin{itemize}

    \item \textit{Is the weather today going to be similar to the one yesterday? And our sun-exposure time? Are your skin and mine similar?}
    \item \textit{Has there been any other day similar to today in which I did happen to need sunscreen? }
\end{itemize}

\caption{Argument from analogy}

\end{subfigure}

\noindent\rule[0.3ex]{\linewidth}{0.4pt}

\vspace{5pt}
\begin{subfigure}{1\linewidth}

\raggedright \textbf{Argument}:

\vspace{5pt}

\centering \textit{"Dr. Smith says sunscreen is not necessary in May, so you don't need it today."} 

\vspace{5pt}

\raggedright \textbf{Critical Questions}:

\begin{itemize}

    \item \textit{ Is Dr. Smith an expert in skin care? Do other experts in skin care agree with Dr. Smith? Is Dr. Smith a trustworthy source? Might Dr. Smith be biased?}
    \item  \textit{ What were the literal words of Dr. Smith? Can his words be checked? }
    \item  \textit{Is his claim consistent with the known evidence about the effects of sun in May?}
\end{itemize}

\caption{Argument from expert opinion}
\end{subfigure}
\noindent\rule[0.5ex]{\linewidth}{1pt}
\caption{Examples of two arguments and the critical questions they raise. While these are synthetic examples, our dataset contains naturally-occurring arguments.}
\label{fig:examplea}
\end{figure}

In this context, researchers have proposed Critical Questions Generation (CQs-Gen) as a method to leverage LLMs to automatically generate critical questions that systematically expose evidential weaknesses or structural flaws (such as fallacies) embedded within argumentative discourse. The ultimate goal is to promote deeper analytical engagement by developing systems capable of generating meaningful critical questions with respect to argumentative texts \cite{calvo-figueras-agerri-2024-critical}.

\textbf{Critical Questions (CQs) are inquiries that may be posed to assess the acceptability of an argument. Therefore, the answer to these questions could potentially challenge the strength of the argument.} 
These questions are closely tied to the specific argumentation schemes underlying the construction of a given discourse \cite{walton_argumentation_2008}. 
As illustrated in Figure \ref{fig:examplea}, an argument can be built based on different schemes, such as \emph{argument from analogy} or \emph{argument from expert opinion}, each triggering a different set of critical questions. Thus, the task of CQs-Gen is formulated as a generative task, in which systems are expected to produce questions that expose the underlying assumptions in the arguments' premises and challenge their inferential structure.

Although interest in these types of tasks is growing \cite{musi_developing_2023, favero2024enhancingcriticalthinkingeducation, ruiz-dolz_detecting_2023}, the lack of datasets and standardized evaluation methods hinders the advancement of research for this particular task. In this work, we introduce the required infrastructure to address these gaps. First, we present the first large-scale dataset including $~$5K manually annotated critical questions. The dataset, collected from multiple sources, includes several critical questions for each text, which are labeled according to a set of validated annotation guidelines. Second, we explore various evaluation strategies for the task, measuring their correlation to human judgments. Finally, we provide a benchmarking by experimenting with 11 LLMs in zero-shot settings, analyzing the diversity of the generated questions, and providing a public leaderboard to encourage further research on CQs-Gen. The main contributions of this work are the following:

\begin{itemize}
    \item The first manually annotated dataset for the task of Critical Questions Generation (CQs-Gen).%
    \item An extensive investigation on the best automatic evaluation methods for this task, establishing that reference-based methods correlate best with human judgments.%
    \item A first benchmarking of the task with 11 state-of-the-art LLMs, which showcases substantial room for improvement, accompanied by a public leaderboard to encourage further research on this particular task.\footnote{\url{https://huggingface.co/spaces/HiTZ/Critical_Questions_Leaderboard}}
\end{itemize}

In the rest of the paper, we first present relevant previous work (Section \ref{sec:previous_work}), explain the dataset creation (Section \ref{sec:dataset}), the CQs-Gen task (Section \ref{sec:task}), and the experimental settings on evaluation and generation (Section \ref{sec:settings}). We then report our experimental findings on evaluation (Section \ref{sec:eval_results}), and use the best evaluation strategies to benchmark 11 LLMs (Section \ref{sec:gen_results}). We finish by analyzing the results in terms of diversity and proposing some future work.

\section{Previous Work}
\label{sec:previous_work}

The automatic generation of critical questions is a relatively new task, introduced and motivated by \citet{calvo-figueras-agerri-2024-critical}. Their work demonstrated that theoretical critical questions, such as those proposed by \citet{walton_argumentation_2008}, and questions generated by LLMs are complementary, as they tend to target different aspects of argumentation. This is in line with the observation that CQs can not be restricted to a prefix set \cite{hernandez_disentangling_2023}. \citet{calvo-figueras-agerri-2024-critical} also identified common failure modes in LLM-generated critical questions, including: (a) producing overly generic or unrelated questions, (b) introducing concepts not present in the source text, (c) flawed reasoning, and (d) generating non-critical questions, such as those resembling reading comprehension tasks.
Parallel to this work, a shared task on Critical Questions Generation has been organized to further stimulate research in this area \cite{calvo2025argminingsharedtask}.

CQs-Gen differs from other question-generation tasks \cite{pan-etal-2020-semantic, miao-etal-2024-discursive} in that its goal is not to produce questions directly answerable from the input text (i.e., reading-comprehension questions). Instead, it aims to elicit questions that uncover what remains unsaid, such as hidden premises or implicit connections.

Moreover, previous work has shown that CQs are useful for fighting misinformation, since they help users identify fallacious reasoning \cite{musi_developing_2023}, and also for predicting the grades of argumentative essays, since they reveal the quality of the arguments \cite{song_applying_2014}. The potential of using CQs in computational applications has been discussed at length \cite{reed_applications_2001, macagno_argumentation_2017, ruiz-dolz_detecting_2023}.

One of the big challenges of generative tasks is evaluation, and CQs-Gen is no different. Thus, machine translation researchers have long been investigating how to use human-generated references to evaluate new text generations using metrics such as BLEURT \cite{sellam-etal-2020-bleurt}, chrF \cite{popovic-2015-chrf}, and COMET  \cite{bosselut-etal-2019-comet}. Similarity metrics such as Semantic Text Similarity (STS) \cite{reimers-2019-sentence-bert} have also been used to compare the new generations to the reference outputs \cite{aynetdinov2024semscoreautomatedevaluationinstructiontuned}. Finally, recent work has shown the effectiveness of using LLMs to evaluate various downstream tasks \cite{zhong2022unified,ke2022ctrleval,jones2024multiaspect,zubiaga-etal-2024-llm}, either by using general-purpose models \cite{wang-etal-2023-chatgpt}, or specialized ones \cite{kim_prometheus_2024, zhu_judgelm_2023}. In this work, we examine all these evaluation strategies to investigate which one is the most suitable for evaluating the CQs-Gen task.

\section{Dataset Creation}
\label{sec:dataset}

In order to benchmark the capacity of current LLMs to generate critical questions, a first step is to create a manually annotated dataset. For this purpose, relevant argumentative texts have been collected, and reference questions have been generated and annotated. The rest of the section provides details of each of these steps.

\subsection{Data Collection}\label{sec:gathering}

To guarantee the relevance of the texts in our dataset, we gather them using four previously available corpora used for argument mining: US2016 \cite{visser_annotating_2021}, Moral Maze Debates (MMD) \cite{lawrence_bbc_2018}, US2016reddit, and Regulation Room Divisiveness (RRD) \cite{konat_corpus_2016}. All these corpora had been annotated with Argumentation Schemes in IAT format \cite{budzynska2011whence}. We reformat these texts by intervention and make the data more manageable by splitting long interventions and merging very short ones. We remove the interventions with no argumentation scheme associated, since these do not necessarily contain any arguments.\footnote{The code to process the IAT diagrams can be found here: \url{https://github.com/hitz-zentroa/critical_questions_generation/tree/main/scripts/pre-process} } %

The data collection process results in a dataset of 220 naturally-occurring interventions, with an average length of 738.4 characters, and 3.1 argumentation schemes. The topics of the interventions range from politics to airline policies and the economy. 

\subsection{References Generation}
\label{sec:ref_gen}

To generate reference CQs for the dataset, we use the method proposed by \citet{calvo-figueras-agerri-2024-critical}. First, we create critical questions based on the templates in \citet{walton_argumentation_2008}. Second, we prompt \textit{Llama-3-70B-Instruct} to write CQs (see Appendix \ref{ann:reference_prompt} for the prompts used). As a result of this process, we obtain an average of 22.4 reference questions per intervention.

\subsection{Annotation}
\label{sec:annotation}

The guidelines for annotating the reference questions have been  
improved in various iterations with the help of the annotators until we reached a moderate inter-annotator agreement (IAA). The annotators were journalists and fact-checkers specialized in detecting misinformation. The annotation process starts with the following question: \textbf{Can this question be used to undermine the arguments given in the intervention?}

If the annotators consider that a question is not useful, they can choose between two options: the question not being valid, or the question being unhelpful. This distinction makes the task more consistent, as the annotators have to be able to describe the reasons why the question is not useful. The three categories are described in the guidelines as follows:\footnote{\url{https://github.com/hitz-zentroa/shared-task-critical-questions-generation/tree/main/shared_task/utils}} 

\begin{enumerate}
    \item \textbf{Useful (USE)}: The answer to this question can potentially challenge one of the arguments in the text.
    \item \textbf{Unhelpful (UN)}: The question is valid, but it is unlikely to challenge any of the arguments in the text.
    \item \textbf{Invalid (IN)}: This question is invalid because it cannot be used to challenge any of the arguments in the text. Either because (1) its reasoning is not right, (2) the question is not related to the text, (3) it introduces new concepts not present in the intervention, (4) it is too general and could be applied to any text, or (5) it is not critical with any argument of the text (e.g. a reading-comprehension question). 
\end{enumerate}

 Additionally, in order to label a question as Unhelpful, the annotators have to provide a short open message stating their reasons. To consider a question Invalid, annotators have to select one of the 5 criteria specified above. 
 This reason-based annotation procedure allowed the annotators to increase their IAA (calculated using Cohen Kappa (Cohen, 1960)) from 0.19 and 0.26 in the first and second pre-annotation rounds (in which we refined the definitions), to 0.54 in the third and final round (in which the reason-based annotation was implemented). In each round, both annotators evaluated 250 questions.

After this, the entire dataset was annotated.
The stats of the annotation per data source can be found in Table \ref{tab:origin_stats}. It can be observed that Useful questions are the most common type of questions. This is especially true in the dataset about airline policies (RRD).

\begin{table}[h]
\centering \scriptsize
\begin{tabular}{r|c|c|ccc}
\toprule
\bf Origin & \bf Nº Int. &  \bf Nº CQs & \bf \% USE &  \bf \% UN & \bf \% IN  \\
\midrule
\bf US2016 & 98 & 2,555 & 59.88 & 23.25 & 16.87 \\
\bf Moral Maze & 25 & 584 & 53.77 & 20.72 & 25.51 \\
\bf RRD & 83 & 1,597 & 66.12 & 23.04 & 10.83 \\
\bf US2016reddit & 14 & 240 & 54.58 & 30.0 & 15.42 \\
\midrule
\bf TOTAL & 220 & 4,976 & 60.91 & 23.21 & 15.88 \\
\bottomrule
\end{tabular}
\caption{Stats of the dataset per source of origin.}
\label{tab:origin_stats}
\end{table}

\subsection{Splitting the dataset}
\label{sec:splits}

Hereafter, we split our dataset between a validation set (publicly available), and a test set. The reference questions of the test set are kept unpublished, since sharing them could result in data contamination that would compromise future evaluations using this dataset \cite{sainz-etal-2023-nlp}. However, we allow future system evaluations through a public leaderboard.

We keep the interventions with the most balanced labels in the test set while putting the rest in the validation set. This splitting procedure increases the quality of the test set, and also makes the reference-based evaluations more accurate (see "Reference-based metrics" in Section \ref{sec:eval_methods}). As observed in Table \ref{tab:stats}, while 67.46\% of the reference questions in the validation set are Useful, this decreases to 42.68\% in the test set.

\begin{table}[h]
\centering \scriptsize
\begin{tabular}{r|c|c|ccc}
\toprule
\bf Set & \bf Nº Int. &  \bf Nº CQs & \bf \% USE &  \bf \% UN & \bf \% IN  \\
\midrule
\bf Validation &   186 & 4,136 &  67.46  &  21.59 &  10.95   \\
\bf Test &  34 &  806 &  42.68 &  31.02 &  26.30  \\
\bottomrule
\end{tabular}
\caption{Stats of the dataset per set.}
\label{tab:stats}
\end{table}

\section{The Task}\label{sec:task}

\begin{figure*}[h!]
\centering \footnotesize 
\begin{minipage}[b]{1\textwidth}
\centering
\begin{subfigure}[b]{\linewidth}
  \centering 
  \begin{tcolorbox}[boxrule=1pt,colback=white,colframe=black] %
  \textit{Walton: Claire's absolutely right about that. 
But then the problem is that that form of capitalism wasn't generating sufficient surpluses. 
And so therefore where did the money flow. 
It didn't flow into those industrial activities, 
 because in the developed world that wasn't making enough money.}
  \end{tcolorbox}
  \vspace{-1em}
  \caption{\textbf{Input}: the intervention}
  \label{ex:implicit}
\end{subfigure}
\end{minipage}%
\hfill %

\vspace{0.2cm} %

\begin{minipage}[b]{.49\textwidth}
\begin{subfigure}[b]{\textwidth}
  \centering \footnotesize
  \begin{tcolorbox}[boxrule=1pt,colback=white,colframe=black]
  \green{USE}: What evidence is there to support the claim that the form of capitalism being used in the developed world was not generating sufficient surpluses? \\ \\
  \green{USE}: How is "sufficient surpluses" defined, and how would one measure it? \\ \\
  \green{USE}: Are there any alternative explanations for why the money did not flow into industrial activities?
  \end{tcolorbox} 
  \vspace{-1em}
  \caption{\textbf{Output}: Given that all CQs here are useful, this answer has an overall punctuation of 1.}
  \label{ex:main}
\end{subfigure}
\end{minipage}
\hfill
\begin{minipage}[b]{.49\textwidth}
\begin{subfigure}[b]{\textwidth}
  \centering \footnotesize
  \begin{tcolorbox}[boxrule=1pt,colback=white,colframe=black]
  \red{IN}: Does this argument support Socialist policies? \\ \\
  \orange{UN}: How does the speaker define "the developed world", and is this a relevant distinction in this context? \\ \\
  \green{USE}: What are the "industrial activities" being referred to, and how do they relate to the form of capitalism in question?
\end{tcolorbox}
  \vspace{-1em}
\caption{\textbf{Output}: This set of questions would get 1/3 points for the useful CQ, 0 for the CQ that is unhelpful, and 0 for the invalid one. Therefore, this answer has a 1/3 punctuation.}
  \label{ex:finance_law_degree}
\end{subfigure}
\end{minipage}
\caption{Example of candidate outputs with its labels: Useful (\green{USE}), Unhelpful (\orange{UN}), and Invalid (\red{IN}).}
\label{fig:main}
\end{figure*}

We formalize the task of CQs-Gen as a generative task where, when given an input argumentative text, the system has to output exactly 3 critical questions. The 3 questions are then evaluated regarding their usefulness for critically assessing the arguments of the text. For the generation of one useful critical question, the task is scored 1/3, for two 2/3, and if the 3 questions are useful, the task is scored with 1, the maximum punctuation. 

In Figure \ref{fig:main}, an instance of the task with two output examples is provided. For this particular intervention, a question such as "\textit{Does this argument support Socialist policies?}" is evaluated as Invalid, as it introduces a new concept which is not present in the text: "Socialist policies". The question "\textit{How does the speaker define "the developed world", and is this a relevant distinction in this context?}" gets evaluated as Unhelpful, as this definition is not central to the arguments of the text. Instead, a similar question such as "\textit{How is "sufficient surpluses" defined, and how would one measure it?}" gets labeled as Useful, as different considerations of what sufficient surpluses are could diminish the strength of the argument.

\section{Experimental Settings}\label{sec:settings}

A major challenge of generative tasks is evaluation. Therefore, before benchmarking different models to generate critical questions, we perform extensive research to find the best evaluation methods. For this purpose, we develop two baseline systems, evaluate their output manually, and assess which evaluation method correlates better with human judgments. The best evaluation methods are then applied to establish the first benchmark with 11 state-of-the-art LLMs. %

\subsection{Evaluation Methods}
\label{sec:eval_methods}

We first perform human evaluation and then assess several automatic approaches: (1) \emph{reference-based metrics}, which compare the newly generated questions to the reference questions, (2) \emph{reference-based LLMs}, which applies LLMs to compare newly generated questions with reference questions, and (3), \emph{labeling LLMs}, which directs LLMs to label the critical questions based on the annotation guidelines. For all these evaluation experiments, we use the output of two baseline systems: \textit{Llama-3-70B-Instruct} and \textit{Qwen2.5-VL-72B-Instruct}, with a very simple prompt (see in Annex \ref{ann:trial_prompt}) in a zero-shot setting. In each step of the evaluation, we aggregate the results of these two baselines, as our goal in this section is not to test the systems but to test the evaluation methods. %

\paragraph{Human evaluation} As a first step, we perform human evaluation of the zero-shot output of the two baseline systems. This manual evaluation allows us to compute statistical correlation with respect to the automatic evaluation methods. The manual judgment consists of visualizing each new question next to the reference questions of that intervention, and selecting the one that inquires about the same information. Then, the new question inherits the label from the reference. If no matching reference is found, we label that question manually following the annotation guidelines (either as Useful, Unhelpful, or Invalid). We validate this evaluation with double annotation and obtain an IAA of 59.14.

\paragraph{Reference-based metrics} We use Semantic Text Similarity (STS) with the Sentence Transformers model \textit{stsb-mpnet-base-v2} \cite{reimers-2019-sentence-bert} to compare each of the newly generated questions to the reference questions in the dataset. Given a threshold, we find the most similar reference question to the new one. If no reference question reaches the threshold, the generated question is given the label \textit{not\_able\_to\_evaluate} (NAE).  Considering $\{R\}$ as the set of vectors of the reference questions, $N$ the vector of the newly generated question, and $T$ the threshold, the label is computed as:

\hspace{1pt}

\noindent
\footnotesize
$f(N) = \begin{cases} 
R_{argmax_{j}cos}(R_j,N) & \text{if } max_{j}cos(R_j,N)>T\\
\text{NAE} & \text{else}
\end{cases}$
\normalsize

\hspace{1pt}

We also experiment with machine translation metrics using this same logic. We use the HuggingFace implementation of BLEURT \cite{sellam-etal-2020-bleurt}, chrF \cite{popovic-2015-chrf}, and COMET \cite{bosselut-etal-2019-comet}.\footnote{\url{https://huggingface.co/evaluate-metric}} For all metrics, the threshold is chosen by comparing the IAA of each threshold with the human evaluation, and the number of values that the threshold left unevaluated (see this comparison in Annex \ref{ann:threshold}).

\paragraph{Reference-based LLMs} As in the previous method, we compare the generated questions to each of the reference questions, and pick the most similar one. However, in this case we apply LLMs to do the matching by querying the models whether both questions ask for the same information. The model can also predict that no reference is similar enough, in other words, the NAE label. The evaluation prompt can be found in Figure \ref{fig:compare_prompt} of Annex \ref{ann:eval_prompt}. As our evaluators, we use \textit{Claude 3.5 Sonnet},\footnote{Version \textit{claude-3-5-sonnet-20241022}, \url{https://www.anthropic.com/news/claude-3-5-sonnet}} a state-of-the-art proprietary closed-weights model, and \textit{Gemma 3 12B Instruct}, an open-weights model. %

\paragraph{Labeling with LLMs} To further explore the potential of LLMs for evaluating the task of CQs-Gen, we now ask the LLM to be the one deciding the evaluation label following the guidelines from Section \ref{sec:annotation}. %
The prompt used for this task for both Claude and Gemma3 is detailed in Figure \ref{fig:zero-shot_prompt} of Annex \ref{ann:eval_prompt}. %
In addition, we also include \textit{Prometheus 2 7B}, an open source LLM specialized in evaluating LLMs \cite{kim_prometheus_2024}. For Prometheus, we adapt the evaluation prompt to reproduce the format it was trained with (see Figure \ref{fig:prometheus-zero} in Annex \ref{ann:eval_prompt}%
). Since Prometheus' instruction includes the requirement of a feedback message explaining why a certain label is given, we also use this feedback to better understand the model's behavior when performing this evaluation (see Annex \ref{ann:pometheus_analysis}).\footnote{We did try changes in the phrasing and complexity of all these prompts (such as asking just for useful vs not-useful questions). We also tried making the LLM choose a reason for its label (as in the annotation process), adding examples of each type of question as in a few-shot approach, and fine-tuning the LLMs for our specific task. However, these changes barely affected the results and, therefore, we do not include them in the paper.}

\paragraph{Augmenting the references.} In a concurrent shared task on CQs-Gen that used this same dataset \cite{calvo2025argminingsharedtask}, a manual evaluation of 15 different submissions was performed. Conveniently, we can now use this data to increase the number of reference questions. In this shared task, each submission generated 3 questions for each intervention in the test set (that is, 45 additional reference questions per intervention). Using these 1,530 ($45*34$) new reference questions, we investigate whether increasing the number of references in the test set improves the results in the reference-based evaluations by incorporating them in batches of 5 per intervention. 

\subsection{Generation Methods}

For benchmarking, we opt for a slightly more elaborate prompt, which includes some additional guidelines (see Annex \ref{ann:gen_prompt}), and apply it to 8 state-of-the-art open-weights models:  \textit{Llama-3-8B-Instruct}, \textit{DeepSeek-R1-Distill-Llama-8B}, \textit{Gemma-2-9b-it}, \textit{Qwen2.5 VL-7B-Instruct},  \textit{Llama-3-70B-Instruct}, \textit{DeepSeek-R1-Distill-Llama-70B}, \textit{Gemma-2-27b-it}, and \textit{Qwen2.5-VL-72B-Instruct},\cite{grattafiori2024llama3herdmodels, deepseekai2025deepseekv3technicalreport, qwen2025qwen25technicalreport, gemmateam2024gemma2improvingopen}; as well as 3 top-performing proprietary models: \textit{Claude 3.5 Sonnet},\footnote{Version \textit{claude-3-5-sonnet-20241022}, \url{https://www.anthropic.com/news/claude-3-5-sonnet}} \textit{GPT-04-mini}, and \textit{GPT-4o} \cite{openai2024gpt4ocard}.\footnote{Versions \textit{gpt-4o-2024-08-06} and \textit{o4-mini-2025-04-16}} Every model is used in their default settings.\footnote{For Gemma2 models, we set the temperature to 0.6, since keeping it at 0 generated the same question 3 times.}

\begin{figure*}[h!]
    \centering
    \includegraphics[width=0.8\linewidth]{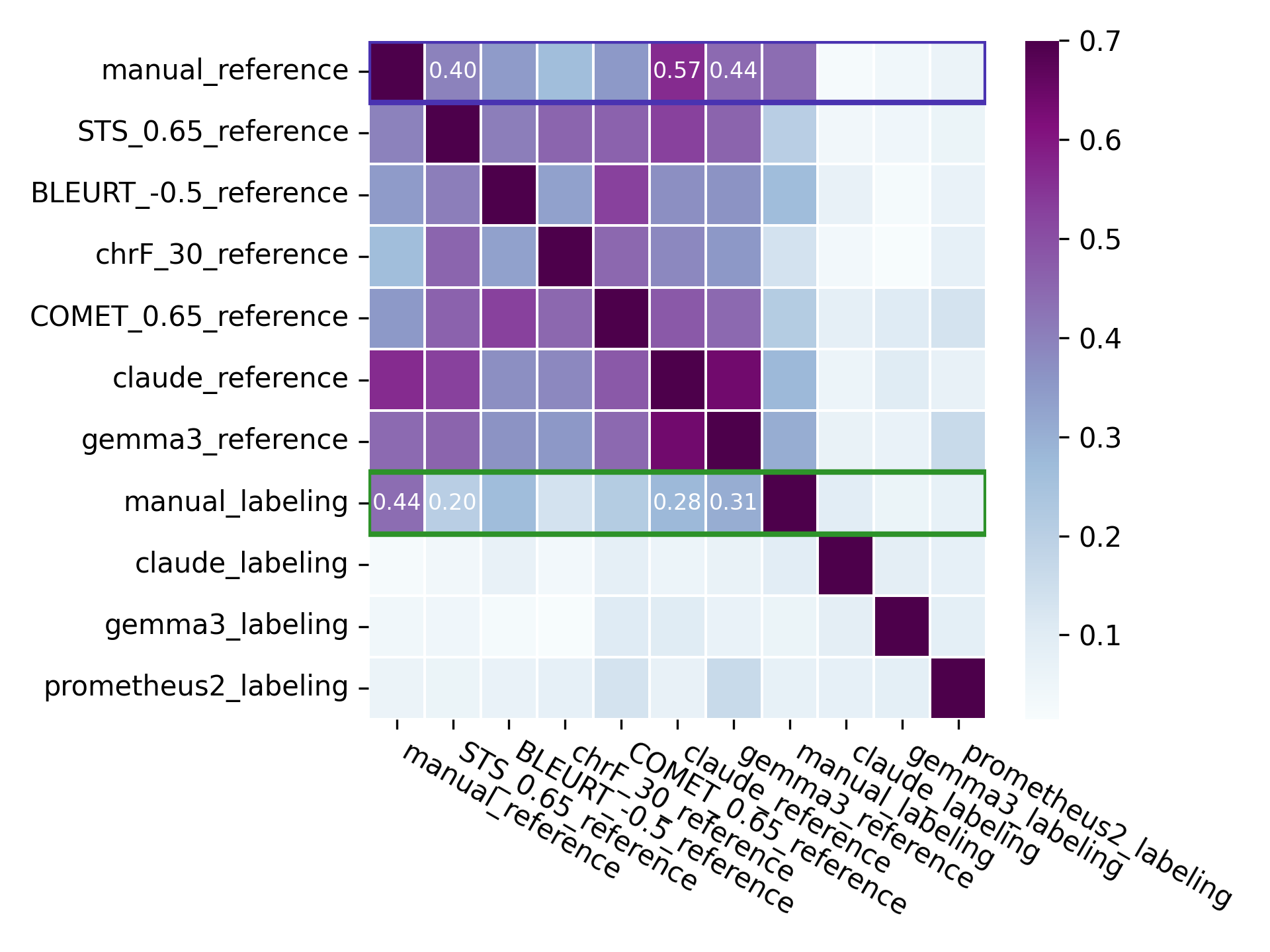}
    \caption{Cohen Kappa scores between all the evaluation methods. These are averaged scores of the two models we manually evaluated: \textit{Llama-3-70B-Instruct} and \textit{Qwen2.5-VL-72B-Instruct}. The numbers in some of the metric names indicate the used threshold.}
    \label{fig:results}
\end{figure*}

\begin{figure*}[h!]
    \centering
    \includegraphics[width=1\linewidth]{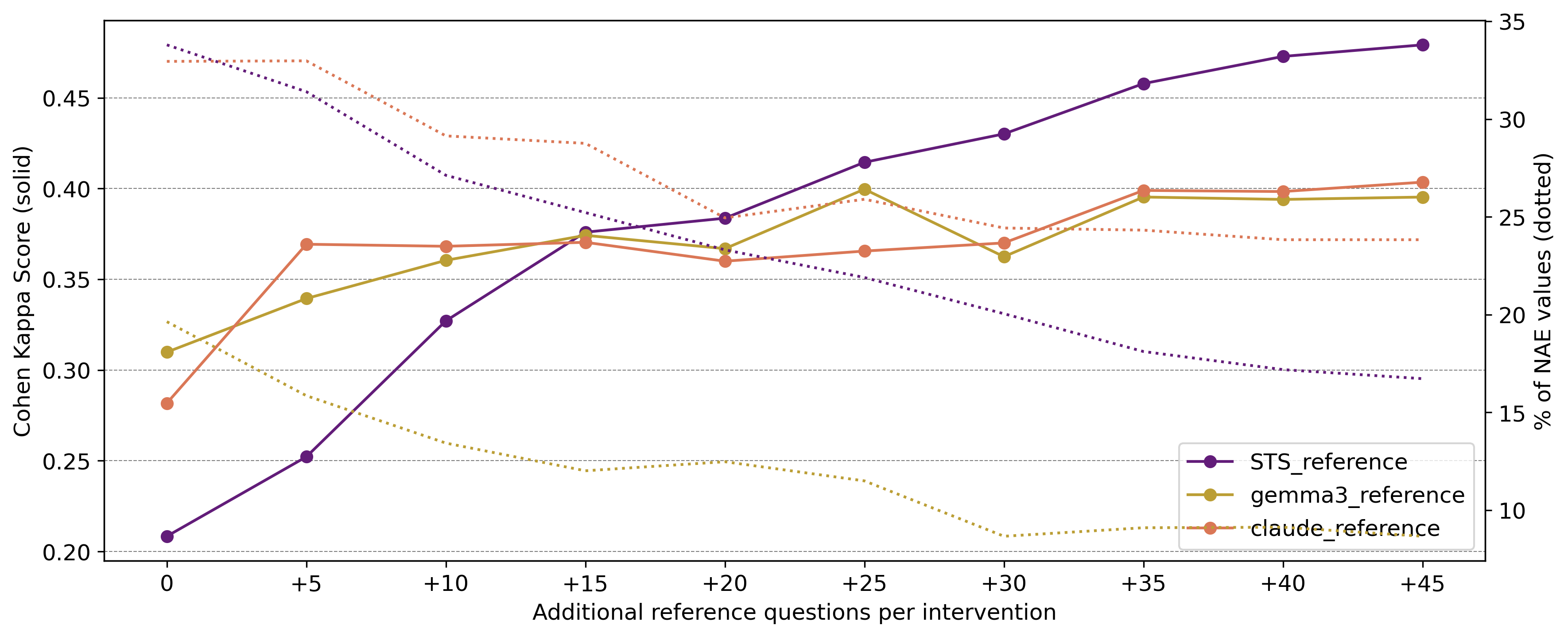}
    \caption{Cohen Kappa Scores between our best automatic evaluation methods and the \textit{manual\_labeling} evaluation when increasing the reference questions in batches of 5 (solid lines). Reported together with the \% of NAE values (dotted lines). These are averaged scores of the two models we manually evaluated: \textit{Llama-3-70B-Instruct} and \textit{Qwen2.5-VL-72B-Instruct}.}
    \label{fig:nae}
\end{figure*}

To evaluate the output of these models, we use the best-performing evaluation methods resulting from the experiments in Section \ref{sec:eval_results}. That is: reference-based STS, reference-based Claude, and reference-based Gemma3. In all cases, we use the test set with the augmented reference questions.

\section{Results}
\label{sec:results}

We first present the results on the various automatic evaluation methods tested, and then we report the performance of the 11 LLMs on the CQs-Gen task.

\subsection{Results on Evaluation Methods}
\label{sec:eval_results}

We compute the results of both of our baseline models (\textit{Llama-3-70B-Instruct} and \textit{Qwen2.5-VL-72B-Instruct}) with each of the evaluation methods defined in Section \ref{sec:settings}. We then compute the agreement between all of them using Cohen Kappa scores (see Figure \ref{fig:results}). We differentiate between the manual evaluation obtained only by matching the newly generated questions with the reference ones and keeping the NAE values (\textit{manual\_reference} in Figure \ref{fig:results}, highlighted in blue), and the setting in which the human evaluator has replaced the NAE values with one of the 3 labels (\textit{manual\_labeling} in Figure \ref{fig:results}, highlighted in green). %

Regarding reference-based methods (both metrics and LLMs), \textit{claude\_reference} obtained the best results, achieving an IAA of 0.57 (moderate agreement) with respect to the \emph{manual\_reference}. \textit{STS\_0.65\_reference} and \textit{gemma3\_reference} also obtain a moderate IAA, with a score of 0.40 and 0.44, respectively (see blue rectangle in Figure \ref{fig:results}).
However, when attempting to evaluate the questions by labeling them directly using LLMs, no evaluation method achieves an IAA higher than 0.1.\footnote{Note that in footnote $^7$ we explain that we tried many other variations of these experiments that did not result in any improvement of the IAA score.} Therefore, from this point onward, we focus only on reference-based evaluation methods. Nonetheless, we provide an error analysis of Prometheus in Annex \ref{ann:pometheus_analysis}, which highlights the difficulties for LLMs to perform this evaluation without relying on gold reference CQs.

While most reference-based methods have a moderate agreement with the \textit{manual\_reference} evaluation, this is not the case when comparing them to the \textit{manual\_labeling} evaluation, where NAE values have been replaced by the right label manually. In this evaluation, STS, Claude, and Gemma3 achieve an IAA of 0.20, 0.28 and 0.31, respectively (see green rectangle in Figure \ref{fig:results}). These results show that, with the current test set, the reference-based evaluation methods are limited by a lack of reference questions.

\begin{table*}[h!]
\footnotesize
    \centering
    \begin{tabular}{r|ccc|c|ccc}
    \toprule
     & \multicolumn{4}{c|}{\bf Score} &  \multicolumn{3}{c}{\bf NAE \%} \\
    \cmidrule(lr){2-5} \cmidrule(lr){6-8}
    method $\longrightarrow$  & \bf STS\_ref. & \bf claude\_ref. & \bf gemma3\_ref. & \bf Average $\downarrow$ & \bf STS & \bf claude & \bf gemma3 \\
    \midrule
    gemma-2-27b & 53.27 $\pm$ 5.7& \underline{51.63} $\pm$ 3.4& 57.19 $\pm$ 2.3 & \bf 54.03 & 11.67 & 16.33 & 3.33 \\

Llama-3-70B & 53.89 $\pm$ 2.6& 48.51 $\pm$ 5.1& \underline{58.15} $\pm$ 6.6 & 53.52 & 4.0 & 14.0 & 1.67 \\
Qwen2.5-72B & \bf \underline{57.73} $\pm$ 1.9& 47.15 $\pm$ 1.5& 53.25 $\pm$ 2.1 & 52.71 & 7.0 & 15.67 & 1.0 \\
gemma-2-9b & 52.94 $\pm$ 3.9& 47.71 $\pm$ 3.4& 55.56 $\pm$ 1.5 & 52.07 & 9.33 & 21.67 & 4.0 \\
DeepSeek-Llama-70B & 47.82 $\pm$ 4.5& 48.73 $\pm$ 1.1& 57.32 $\pm$ 2.7 & 51.29 & 16.33 & 10.67 & 1.0 \\
Llama-3-8B & 50.24 $\pm$ 3.9& 42.34 $\pm$ 4.2& 55.12 $\pm$ 3.9 & 49.23 & 11.33 & 21.33 & 3.0 \\
DeepSeek-Llama-8B & 44.88 $\pm$ 3.9& 36.18 $\pm$ 1.8& 50.73 $\pm$ 5.5 & 43.93 & 16.0 & 27.67 & 4.33 \\

Qwen2.5-7B & 43.30 $\pm$ 1.4& 43.94 $\pm$ 1.3& 42.97 $\pm$ 1.2 & 43.40 & 6.33 & 12.0 & 1.33 \\

\midrule
claude-3-5-sonnet & 56.21 $\pm$ 0.6& 50.33 $\pm$ 1.5& \bf 62.42 $\pm$ 1.5 & \bf 56.32 & 9.33 & 16.0 & 3.33 \\
gpt-4o & 52.29 $\pm$ 1.5& \bf 52.94 $\pm$ 2.9& 58.17 $\pm$ 3.4 & 54.47 & 13.0 & 14.33 & 1.0 \\
o4-mini & 53.27 $\pm$ 6.0& 50.33 $\pm$ 4.9& 58.82 $\pm$ 2.6 & 54.14 & 10.0 & 12.0 & 1.33 \\

\bottomrule
    \end{tabular}
    \caption{Generation results obtained with the prompt in Annex \ref{ann:gen_prompt}. In the reference-based evaluation, the \textit{extended\_test} was used, and NAE values were considered not-useful. We report average of 3 runs and standard deviation. \textbf{Bold}: best overall results per metric; \underline{underlined}: best open-weights results per metric. All models are instruct models.}
    \label{tab:generation_results}
\end{table*}

\noindent \textbf{Extending the test set:} To address the scarcity of reference questions, we extend our test set by incorporating the 15 manually evaluated submissions from the CQs-Gen Shared Task as additional references. That is 45 extra reference questions per intervention (which amounts to 68.64 per intervention on average). In Figure \ref{fig:nae}, we incorporate these new references in batches of 5, and calculate the resulting IAA and percentage of NAE values. As observed, the inclusion of new references leads to a higher IAA. For \textit{STS\_reference}, this improvement is steady as more references are added, reaching an IAA of 0.48 when all 45 questions are added. In contrast, for LLM-based reference methods, the gains level off quickly when reaching an IAA of 0.40. This divergence arises from methodological factors: while STS evaluates each reference question independently, LLM-based methods integrate all reference questions into the prompt, causing the context length to increase with each addition. %
However, since the quality of the evaluation does not decrease either, we will be using all the references in the upcoming evaluations. It should also be noted that \textit{claude\_reference} is more strict than \textit{gemma3\_reference} when evaluating, leaving a higher percentage of questions unevaluated.

In order to test if these three automatic evaluation methods are suitable for comparing CQs-Gen systems, we now use this method to re-evaluate all the submissions of the shared task. For a fair evaluation, we remove the references belonging to the submission being evaluated at each time. With this experiment, we observe that the ranking generated by Claude and Gemma correlates to the humanly evaluated ranking of the shared task with a Person's Coefficient of 0.86 and 0.80, respectively. For \textit{STS\_reference}, the correlation is 0.69.

Therefore, each of these three evaluation methods presents distinct strengths and limitations. The \textit{claude\_reference} approach achieves the highest correlation with the shared task ranking, yet it leaves a substantial number of instances unevaluated. In contrast, \textit{gemma3\_reference} minimizes the number of unevaluated cases, but it exhibits the lowest IAA with human annotations. Finally, \textit{STS\_reference} achieves the strongest IAA with human annotations, but it performs the weakest in terms of model comparison.

Considering these results, our benchmarking will report these three metrics alongside the percentage of NAE values, which serves as a proxy for the confidence of the evaluation. For the test set, we will use all the references available (from now on, \textit{extended\_test}).

\subsection{Results on CQs-Gen}\label{sec:gen_results}

Table \ref{tab:generation_results} presents the benchmarking results of 11 state-of-the-art LLMs (generation prompt in Annex \ref{ann:gen_prompt}), evaluated using our three strongest evaluation methods. The results indicate that, in a zero-shot setting, no model generates more than 57\% of Useful CQs on average. It is important to note, however, that no parameter optimization or prompt tuning has been applied in this evaluation. Consequently, higher scores are reported in \citet{calvo2025argminingsharedtask}, where systems specifically designed for this task achieved results nearly 10 points higher.

For the open-weight models, the best-performing LLM is \textit{gemma-2-27b-it}, which achieves an average score of 54.03, slightly higher than \emph{Llama-3-70B-it}, and comparable to the closed-weight proprietary models. We also observe consistent, though modest, improvements when using larger variants within each model family. The highest overall performance is achieved by \emph{claude-3-5-sonnet}.

However, the different evaluation methods differ in their judgments, with \textit{claude\_reference} making \textit{gpt-4o} the winner, \textit{gemma3\_reference} making \textit{claude-3-5-sonnet}, and \textit{STS\_reference} making \textit{Qwen2.5-VL-72B}. These differences should be further studied. In all cases, \textit{gemma3\_reference} has the lowest percentage of NAE values, and \textit{claude\_reference} the highest.

\section{Qualitative Analysis}
\label{sec:analysis}

While the primary goal of this task is to generate useful critical questions, the diversity of these questions also serves as an important indicator of quality. A broader range of questions may reflect a deeper comprehension of the texts, while a minimal-effort approach characterized by the repeated use of identical or formulaic templates may suggest a superficial understanding and limited capacity for critical engagement.

To investigate this issue, we use two metrics, namely, n-gram diversity score and compression ratio diversity score (CR-div) \cite{reimers-2019-sentence-bert}. The n-gram metric calculates diversity as the ratio of the number of unique n-grams to the total n-grams occurring in the entire set of generated questions. The compression ratio (CR) comes from calculating the ratio of the size of the compressed file to its original size. The CR diversity score can be calculated as the reciprocal of the compression ratio to get a score between 0 and 1, 0 being for identical documents. \citet{shaib2025standardizingmeasurementtextdiversity} shows that this metric is the most reliable one.

Table \ref{tab:diversity} shows that both Qwen2.5 models produce the least diverse output, closely followed by \textit{claude-3-5-sonnet}. These results also hold when we look only at Useful CQs (third column of Table \ref{tab:diversity}). The most diverse output is produced by \textit{o4-mini}. Taking a closer look at the output of Qwen2.5-VL models, we observe that half of the CQs generated by these models follow one of these templates:

\begin{table}[h!]
\footnotesize
    \centering
    \begin{tabular}{r|ccc}
    \toprule
  metric $\longrightarrow$  & \bf n-gram & \bf CR-div $\downarrow$  & \bf  CR-d(USE) \\
    \midrule
    DeepSeek-70B & 2.71 & \bf .313 & .339 \\
    DeepSeek-8B & 2.65 & .309 & .335 \\
    Llama-3-70B & 2.52 & .293 & .335 \\
Llama-3-8B & 2.41 & .289 & .323 \\
gemma-2-9b & 2.45 & .287 & .304 \\
gemma-2-27b & 2.45 & .282 & .300 \\
Qwen2.5-72B & 1.04 & .\underline{131} & .146 \\
Qwen2.5-7B & 0.97 & \underline{.123} & .140 \\
\midrule
o4-mini & 2.76 & \bf .332 &  .351 \\
gpt-4o & 2.77 &  .313 &  .336 \\
claude-3-5-sonnet & 1.36 & \underline{.167} & .183 \\
\bottomrule
    \end{tabular}
    \caption{Diversity metrics calculated within the CQs generated by models in Table \ref{tab:generation_results}. \textit{CR-d(USE)} shows the diversity of Useful CQs. \textbf{Bold}: best overall results per metric; \underline{underlined}: worst overall results per metric. }
    \label{tab:diversity}
\end{table}

\begin{itemize}
    \item How does \{speaker\} address \{related\_matter\}?
    \item What evidence does \{speaker\} provide to support the claim that \{claim\}?
\end{itemize}

In the same line, \textit{claude-3-5-sonnet} also outputs a recursive template regarding evidence, mainly:

\begin{itemize}
    \item  What evidence supports the claim that \{claim\}?
\end{itemize}

However, both the 70B Qwen model and Claude perform substantially better at selecting claims to populate the template than the smaller Qwen model, leading to a higher proportion of Useful critical questions (see Table \ref{tab:generation_results}). As highlighted in the CQs-Gen shared task, diversity is a relevant metric for this task, as it serves as an important indicator of the model’s level of understanding, which can not be observed from the overall score alone.

\section{Conclusion and Future Work}

In this work, we have introduced a comprehensive infrastructure for the development and evaluation of systems addressing the task of Critical Questions Generation. First, we constructed a dataset compiled from four different sources, annotated with critical questions categorized as Useful, Unhelpful, or Invalid. Second, we explored automatic evaluation methods for this task and found that reference-based approaches 
yield the most reliable results. Finally, we conducted zero-shot evaluations across 11 LLMs, revealing substantial room for improvement. To support future development, we also provide a public leaderboard for benchmarking CQs-Gen systems.

Although the dataset includes a validation and a test set, our results are based solely on the test set, as we did not pursue parameter or prompt tuning to improve model performance.
Consequently, exploring methods for improving model performance represents a natural direction for future work.

In addition, we aim to investigate the practical utility of generating critical questions for both automated systems and human users. For automated systems, a promising direction is to assess whether incorporating critical questions as intermediate steps in complex reasoning tasks can enhance overall performance. For human users, it could be explored whether exposure to (automatically generated) critical questions could foster improved critical thinking skills.

\section*{Limitations}

Despite the contributions of this work, several limitations remain. First, our dataset, while of reasonable size, is certainly extendable, although this is hindered by the scarcity of datasets annotated with argumentation schemes, particularly beyond English. Thus, expansion using only LLM-generated critical questions remains a viable path even though the annotation should be performed by experts, which makes it an expensive effort. 
Second, although we have reduced the number of unevaluated questions in the test set, this shortcoming remains an open research question.

\section*{Acknowledgments}

We are thankful to the following MCIN/AEI/10.13039/501100011033 projects: (i) DeepKnowledge (PID2021-127777OB-C21) and by FEDER, EU; (ii) Disargue (TED2021-130810B-C21) and European Union NextGenerationEU/PRTR; (iii) DeepMinor (CNS2023-144375) and European Union NextGenerationEU/PRTR. Blanca Calvo Figueras is supported by the UPV/EHU PIF22/84 predoc grant. We would like to acknowledge the contributions of Celia Ramos and Pablo Hernández from Maldita.es for annotating the dataset and giving feedback on the guidelines, as well as Maite Heredia, for performing the secondary annotations.
\bibliography{custom,Argument-disinformation}

\clearpage

\onecolumn

\appendix

\section{Reference Generation Prompts}
\label{ann:reference_prompt}

\begin{figure}[h]
\centering \footnotesize
\begin{minipage}[b]{1\textwidth}
  \centering 
  \begin{tcolorbox}[boxrule=1pt,colback=white,colframe=mycustomcolor] %

List the critical questions that should be asked regarding the arguments in the following paragraph:

\{intervention\}

  \end{tcolorbox}
  \vspace{-1em}
  \begin{tcolorbox}[boxrule=1pt,colback=white,colframe=mycustomcolor] %

Suggest which critical questions should be raised before accepting the arguments in this text:

\{intervention\}

  \end{tcolorbox}
   \vspace{-1em}
  \caption{Prompt for generating the reference questions.}
  \label{fig:reference_prompt}
\end{minipage}
\end{figure}

\section{Baselines Prompt for Evaluation Experiments}
\label{ann:trial_prompt}

\begin{figure}[h]
\centering \footnotesize
\begin{minipage}[b]{1\textwidth}
  \centering 
  \begin{tcolorbox}[boxrule=1pt,colback=white,colframe=mycustomcolor] %

Give me 3 or more critical questions that should be raised before accepting the arguments in this text:

\{intervention\}
                
Give one question per line. Make sure there are at least 3 questions. Do not give any explanation regarding why the question is relevant.

  \end{tcolorbox}
   \vspace{-1em}
  \caption{Prompt for generating the output of the baselines.}
  \label{fig:trial_prompt}
\end{minipage}
\end{figure}

\section{Threshold selection for reference-based metrics}
\label{ann:threshold}

\begin{figure}[h!]
    \centering
    \includegraphics[width=1\linewidth]{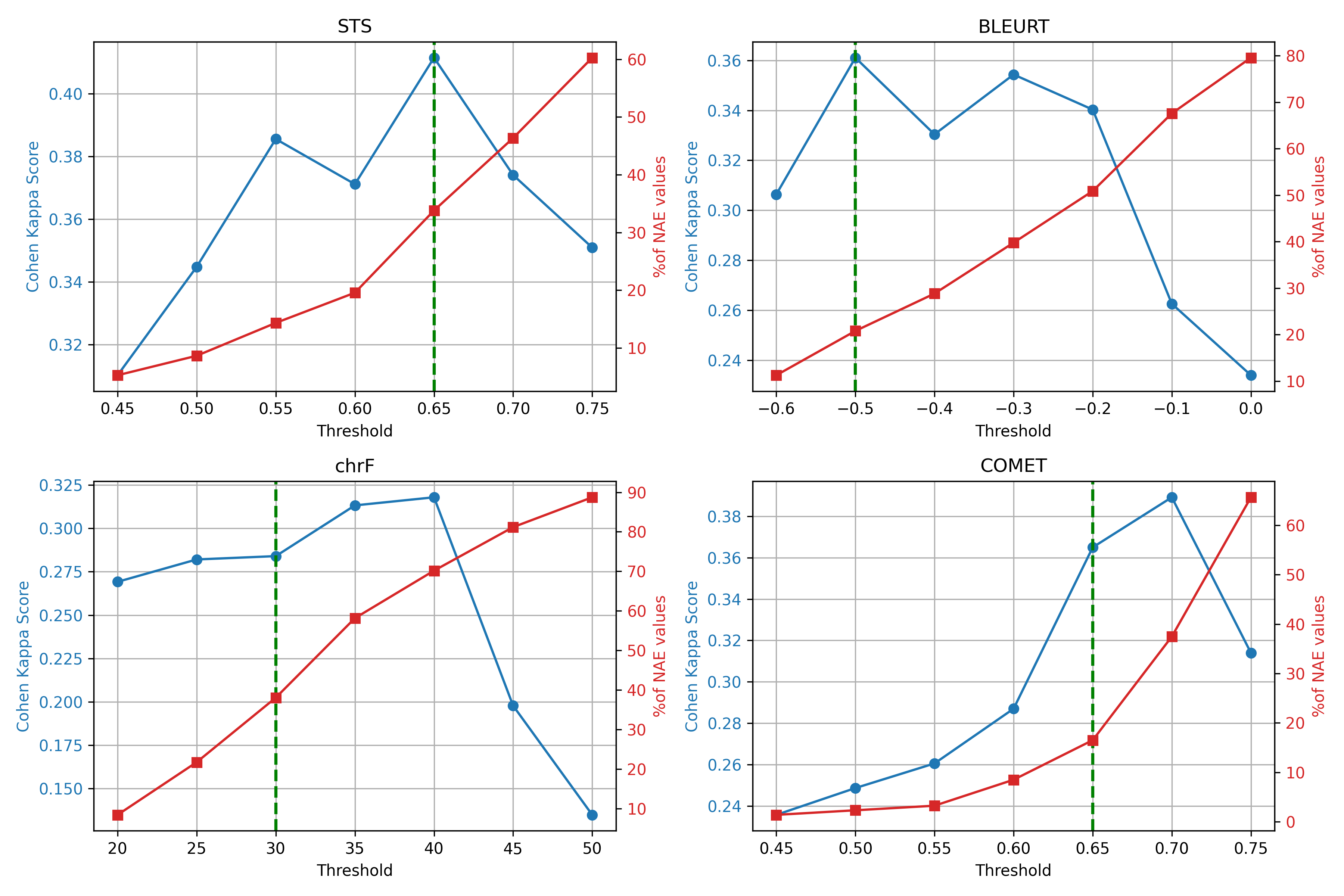}
    \caption{The threshold was selected to give the best balance between the Cohen Kappa Score (blue dots), and the \% of NAE values it generated (red squares). The chosen threshold for each metric is highlighted in green.}
    \label{fig:threshold}
\end{figure}

\clearpage

\section{Evaluation Prompts}
\label{ann:eval_prompt}

\begin{figure}[h]
\footnotesize 
\begin{minipage}[b]{1\textwidth}
  \centering 
  \begin{tcolorbox}[boxrule=1pt,colback=white,colframe=mycustomcolor] %

You will be given a set of reference questions, each with an identifying ID, and a newly generated question. Your task is to determine if any of the reference questions are asking for the same information as the new question.
        
Here is the set of reference questions with their IDs:

<reference\_questions>

\{references\}

</reference\_questions>

Here is the newly generated question:

<new\_question>

\{cq\}

</new\_question>

Compare the new question to each of the reference questions. Look for questions that are asking for the same information, even if they are worded differently. Consider the core meaning and intent of each question, not just the exact wording.

If you find a reference question that is asking for the same information as the new question, output only the ID of that reference question.

If none of the reference questions are asking for the same information as the new question, output exactly 'Similar reference not found.' (without quotes).

Your final output should consist of only one of the following:

1. The ID of the most similar reference question

2. The exact phrase 'Similar reference not found.'

Do not include any explanation, reasoning, or additional text in your output.

  \end{tcolorbox}
  \vspace{-1em}
  \caption{Prompt for comparing the newly generated questions to the reference questions. }
  \label{fig:compare_prompt}
\end{minipage}

\end{figure}

\begin{figure}[t!]

\begin{minipage}[b]{1\textwidth}
  \centering \scriptsize
  \begin{tcolorbox}[boxrule=1pt,colback=white,colframe=mycustomcolor] %

 You are a fair judge assistant tasked with evaluating if a provided question is a useful critical question for a given text. Your role is to provide clear objective feedback based on specific criteria, ensuring each assessment reflects the absolute standards set for performance.

Here is the question you should evaluate:

<critical\_question>

\{cq\}

</critical\_question>

And here is the text to which the question relates:

<text>

\{intervention\}

</text>

Guidelines for evaluation:

1. Carefully read both the question and the text.

2. Consider how the question relates to the arguments presented in the text.

3. Assess the question's usefulness in challenging or critically examining the text's content.

4. Determine which of the three labels (Useful, Unhelpful, or Invalid) best applies to the question.

Label criteria:

1. Useful: The question is both critical of and directly relevant to the arguments in the text. It challenges the text's content in a meaningful way.

2. Unhelpful: The question is critical and related to the text, but not likely to be very useful in challenging its arguments. This could be because:

   a) The answer is common sense
   
   b) The answer is well-known and not controversial
   
   c) The question is very complicated to understand or answer
   
   d) The text itself already answers the question
   
   Note: Do not use this label just because better questions could have been posed.

3. Invalid: The question is not appropriately critical in this context. This could be because:

   a) The question is unrelated to the text
   
   b) The question is too general and could apply to many texts
   
   c) The question introduces new concepts not mentioned in the text
   
   d) The question doesn't challenge any arguments in the text (e.g., it's a simple reading comprehension question or asks about the speaker's/reader's opinion)
   
   e) The question critiques an argument that the speaker wasn't actually making

Your task is to output only one of the three labels: Useful, Unhelpful, or Invalid. Do not include any comments, explanations, blank spaces, or new lines. Your entire output should consist of a single word - the chosen label.

  \end{tcolorbox}
  \vspace{-1em}
  \caption{Prompt for directly labeling the newly generated questions using Claude and Gemma3. }
  \label{fig:zero-shot_prompt}
\end{minipage}
\end{figure}

\begin{figure}
\centering \scriptsize
\begin{minipage}[b]{1\textwidth}
  \centering 
  \begin{tcolorbox}[boxrule=1pt,colback=white,colframe=mycustomcolor] %

You are a fair judge assistant tasked with providing clear, objective feedback based on specific criteria, ensuring each assessment reflects the absolute standards set for performance.
        
\#\#\#Task Description:

An instruction (might include an Input inside it), a response to evaluate, some reference answers for score 1, 2 and 3, and a score rubric representing a evaluation criteria are given.

1. Write a detailed feedback that assess the quality of the response strictly based on the given score rubric, not evaluating in general.

2. After writing a feedback, write a score that is an integer between 1 and 3. You should refer to the score rubric.

3. The output format should look as follows: ``Feedback: (write a feedback for criteria) [RESULT] (an integer number between 1 and 3)''

4. Please do not generate any other opening, closing, and explanations.

\#\#\#The instruction to evaluate:

1. Carefully read both the question and the text.

2. Consider how the question relates to the arguments presented in the text.

3. Assess the question's usefulness in challenging or critically examining the text's content.

4. Determine which of the three labels (Useful, Unhelpful, or Invalid) best applies to the question.

Here is the question you should evaluate:

<critical\_question>

\{cq\}

</critical\_question>

And here is the text to which the question relates:

<text>

\{intervention\}

</text>

\#\#\#Score Rubrics:

[Is the question posed by the model critical and useful for challenging the arguments in the text?]

Score 1: Invalid--> The question is not appropriately critical in this context. This could be because:

   a) The question is unrelated to the text
   
   b) The question is too general and could apply to many texts
   
   c) The question introduces new concepts not mentioned in the text
   
   d) The question doesn't challenge any arguments in the text (e.g., it's a simple reading comprehension question or asks about the speaker's/reader's opinion)
   
   e) The question critiques an argument that the speaker wasn't actually making
   
Score 2: Unhelpful--> The question is critical and related to the text, but not likely to be very useful in challenging its arguments. This could be because:

   a) The answer is common sense
   
   b) The answer is well-known and not controversial
   
   c) The question is very complicated to understand or answer
   
   d) The text itself already answers the question
   
   Note: Do not use this label just because better questions could have been posed.
   
Score 3: Useful --> The question is both critical of and directly relevant to the arguments in the text. It challenges the text's content in a meaningful way.

\#\#\#Feedback:

  \end{tcolorbox}
  \vspace{-1em}
  \caption{Prompt for directly labeling the newly generated questions using Prometheus 2 7B. }
  \label{fig:prometheus-zero}
\end{minipage}
\end{figure}

\clearpage

\section{Analysis of Prometheus' evaluation}
\label{ann:pometheus_analysis}

\footnotesize

Using the prompt in Figure \ref{fig:prometheus-zero} and Prometheus 2 7B \cite{kim_prometheus_2024},  we tried to evaluate both baselines (\textit{Llama-3-70b} and \textit{Qwen-2.5-72b}). The results in Figure \ref{fig:results} show a very low IAA with respect to human evaluation. To further investigate this issue we inspect the errors, observing that 25\% of the questions get labeled as Useful when they should have been labeled as Invalid, 23.5\% of the questions get labeled as Useful when the annotators had labeled them as Unhelpful, and 16.1\% are judged as Unhelpful instead of Useful. 

In the table below, we show some examples of these 3 error directions. In many instances, if the generation model had misunderstood the text when generating the question (bad reasoning error), the judge model just followed along with the model's interpretation (as in the first example below). In other cases, the judge model tried too hard to find a reason for the question to be critical, as in example 2. In other instances, the judge model failed to interpret the implicit relations that the speaker was drawing (as in example 4). In the third column below, you can find the explanation of each of these errors.

\begin{table}[ht]
    \centering \footnotesize 
    \begin{tabular}{m{5cm}m{6cm}m{4cm}}
    \toprule 
        \bf Intervention, Question 
        
        and Gold Label & \bf   Judge Feedback and Label & \bf  Error in the Labeling \\ 
        \midrule 

\textbf{Intervention}: TRUMP: "I did not support the war in Iraq

That is a mainstream media nonsense put out by her

The best person in her campaign is mainstream media"

\hfill

\textbf{Question}: Can you identify examples of how the "best person in her campaign" has been influenced by mainstream media?

\hfill

\textbf{Gold Label}: Invalid &

\textbf{Feedback}:  In the provided text, statement by statement, it can be seen that the question asked about examples of the candidate being influenced by mainstream media. \textbf{The text, however, directly states that mainstream media is a "nonsense" put out by the best person in the campaign}. This statement, which is clearly within the context of the text, directly addresses the question posed. Therefore, it does challenge the arguments presented in the text and adds an informative layer about the speaker's influence by mainstream media. [...]

\hfill

\textbf{Judge Label}: Useful & 
In the text, Trump is saying that mainstream media have been on the side of his opponent. Basically, he is claiming that media are biased towards her and are disseminating lies about him. However, the question does not understand his point, instead, it assumes it is someone in particular from her campaign the one who has been "influenced by mainstream media". The judge falls into the same trap.\\
\midrule
\textbf{Intervention}: HOLT: "How do you heal the divide

The share of Americans who say race relations are bad in this country is the highest it's been in decades

much of it amplified by shootings of African-Americans by police we've seen recently in Charlotte and Tulsa

Race has been a big issue in this campaign

one of you is going to have to bridge a very wide and bitter gap"

\hfill

\textbf{Question}: Are there historical precedents for successfully bridging such divides, and if so, what were they?

\hfill

\textbf{Gold Label}: Invalid & 

\textbf{Feedback}:  Based on the score rubric, the provided question aligns with the criteria of applying critical analysis to the arguments presented in the text. The question requires a thoughtful consideration of historical precedents that might have been successful in "healing the divide" that the speaker alludes to. \textbf{It directly challenges the implicit assumption in the text that it's necessary to seek a solution or to address the issue of race relations.} The question also refrains from introducing new or unrelated concepts, which is a criterion specified in the rubric. [...] %

\hfill

\textbf{Judge Label}: Useful & 
The judge model claims that the existence of "historical precedents" challenges the idea that there is a need for solutions. However, this is not true, as solutions might be needed even if other solutions have been implemented in the past.

The annotator labeled the question as invalid because it does not challenge any of the arguments made by the speaker. On the contrary, the question could have come from Holt himself. \\
\midrule

\textbf{Intervention}: HOLT: "How do you heal the divide

The share of Americans who say race relations are bad in this country is the highest it's been in decades

much of it amplified by shootings of African-Americans by police we've seen recently in Charlotte and Tulsa

Race has been a big issue in this campaign

one of you is going to have to bridge a very wide and bitter gap"

\hfill

\textbf{Question}: Is the statement about the share of Americans who say race relations are bad supported by data?

\hfill

\textbf{Gold Label}: Useful & 

\textbf{Feedback}: The response to the question shows an understanding of the query, as it assesses the level of data support for the statement made in the text. However, the response doesn't dive deep into how the question relates to the arguments presented in the text, and it fails to offer a critique of these arguments. This lack of critical analysis, therefore, limits its usefulness for challenging the text's content. Moreover, \textbf{the question posed is somewhat vague and doesn't necessarily probe into all aspects of the statement.} There's room for improvement in demonstrating the criticality of the question and in examining its utility for challenging the arguments within the text. [...] "

\hfill

\textbf{Judge Label}: Unhelpful & 

The main argument of this text is that racial relations are getting worse in America. To argue that, he uses several premises, and one of these premises is that the share of the Americans who say so, has gone up. Therefore, not having data to sustain this claim would diminish the validity of his argument. The judge dismisses the question saying that it does not "probe into all aspect of the argument". However, this is not a requirement in the rubric.\\

\midrule
    \end{tabular}
\end{table}

\begin{table}[h!]
    \centering \footnotesize 
    \begin{tabular}{m{5cm}m{6cm}m{4cm}}
    \hline 
        \bf \footnotesize  Intervention, Question 
        
        and Gold Label & \bf \footnotesize  Judge Feedback and Label & \bf \footnotesize Error in the Labeling \\ 
        \midrule 

\textbf{Intervention}: CLINTON: "actually, I have thought about this quite a bit

not quite that long my husband did a pretty good job in the 1990s

I think a lot about what worked and how we can make it work again

a balanced budget, million new jobs

Incomes went up for everybody, Manufacturing jobs went up also in the 1990s, if we're actually going to look at the facts [...]"

\hfill

\textbf{Question}: How do we know that Clinton's husband's policies were the cause of the economic growth in the 1990s?

\hfill

\textbf{Gold Label}: Useful & 

\textbf{Feedback}: The question \textbf{seems to assume a causal relationship between the policies of Bill Clinton and the economic growth of the 1990s, which is not explicitly stated in the text and could introduce ambiguity.} It asks for evidence to support a statement that is not clearly stated as part of the text's content. This could potentially lead to misunderstanding or misinterpretation of the text's actual stance. Additionally, the question appears to critique an argument that the speaker was not actually making, which could be considered invalid in terms of critical analysis. It does, however, stay within the confines of the context, only dealing with policies and economic growth of the 1990s. The question, although critical, lacks the useful edge that it should have to truly challenge the text's arguments. This gives it a certain level of unhelpfulness. [...] %

\hfill

\textbf{Judge Label}: Unhelpful & 
While the relation between Bill Clinton's policies and the economic growth in the 1990s might not be strictly stated in the text, the annotators interpreted that Hillary is implying that this relation exists as, otherwise, her mentioning these two facts together would be pointless. Therefore, if we interpret Hillary is drawing this relation, it makes a lot of sense to ask if it is actually a correct one.\\

\midrule

\textbf{Intervention}: CLINTON: "that is just not accurate

I was against it once it was finally negotiated and the terms were laid out"

\hfill

\textbf{Question}: Who negotiated the terms?

\hfill

\textbf{Gold Label}: Unhelpful & 

\textbf{Feedback}:  The question "Who negotiated the terms?" is directly relevant to the text provided. It asks for specific information that is critical in understanding the context and negotiation dynamics, which were clearly stated by the speaker, Clinton. The question doesn't divert into new concepts but centers on the question of agency - in this case, the role of the negotiator - which is a fundamental aspect of negotiations. This aligns with the goal of challenging the content of the text, as specified in the rubric. [...] %

\hfill

\textbf{Judge Label}: Useful & 
While who negotiated the terms could potentially decrease the validity of the claim (for instance, in case it was actually her who did it), it is unlikely to do so. \\
\midrule
\textbf{Intervention}: CLINTON: "I support our democracy

sometimes you win, sometimes you lose

I certainly will support the outcome of this election

Donald's trying very hard to plant doubts about it

I hope the people out there understand this election's really up to you

It's not about us so much as it is about you and your families and the kind of country and future you want

I sure hope you will get out and vote as though your future depended on it

it does"

\hfill %

\textbf{Question}: Is Clinton's appeal to people to vote based on their own interests rather than the common good?

\hfill

\textbf{Gold Label}: Unhelpful & 

\textbf{Feedback}:  The critical question posed here directly engages with the arguments provided in the text. It focuses on a specific aspect of the argument – the emphasis on voting based on personal interests versus the common good. While it is critical and directly related to the arguments made by Clinton, it lacks depth and could be potentially well-known to some readers, thus not entirely challenging. The question, though pertinent, may fall short of providing substantial insight or critiquing the text's content in a more nuanced manner. However, it still manages to be useful in challenging the arguments presented by the speaker, as it specifically targets the key issue within the text. Therefore, according to the score rubric, this question is useful in challenging the arguments in the text. 

\hfill

\textbf{Judge Label}: Useful & 
Clinton's argument in this text is that people should go to vote because it's their future what is at play. While it is true that she might have selfish interests to be encouraging people to vote, that is unlikely to diminish the validity of her argument.\\
\bottomrule
    \end{tabular}
    \caption{Examples of the feedback given by Prometheus 2 for the most typical kinds of errors. The reasons why the label was not correct are given by the authors of this paper on the last column. Some feedback texts and interventions were shortened in order to fit this table. }
\end{table}

\clearpage

\section{Generation Prompt}
\label{ann:gen_prompt}

\begin{figure}[h]
\centering \footnotesize
\begin{minipage}[b]{1\textwidth}
  \centering 
  \begin{tcolorbox}[boxrule=1pt,colback=white,colframe=mycustomcolor] %

You are tasked with generating critical questions that are useful for diminishing the acceptability of the arguments in the following text:
                
\{intervention\}
                
Take into account a question is not a useful critical question:

- If the question is not related to the text.

- If the question is not specific (for instance, if it's a general question that could be applied to a lot of texts).

- If the question introduces new concepts not mentioned in the text (for instance, if it suggests possible answers).

- If the question is not useful to diminish the acceptability of any argument. For instance, if it's a reading-comprehension question or if it asks about the opinion of the speaker/reader.

- If its answer is not likely to invalidate any of the arguments in the text. This can be because the answer to the question is common sense, or because the text itself answers the question.
                
Output 3 critical questions.        

Give one question per line. 

Make sure there are at least 3 questions.

Do not give any other output.

Do not explain why the questions are relevant.

  \end{tcolorbox}

   \vspace{-1em}
  \caption{Generation Prompt for benchmarking.}
  \label{fig:gen_prompt}
\end{minipage}
\end{figure}

\end{document}